\documentclass[manuscript]{acmart}

\usepackage{microtype}
\usepackage{graphicx}
\usepackage{subfigure}
\usepackage{booktabs} 
\usepackage{tabularx}
\usepackage{multirow}  
\usepackage{multicol}

\usepackage{mathtools}
\usepackage{amsmath} 

\usepackage{algorithm}
\usepackage{algorithmic}

\copyrightyear{2022} 
\acmYear{2022} 
\setcopyright{acmlicensed}\acmConference[IVSP 2022]{2022 4th International Conference on Image, Video and Signal Processing}{March 18--20, 2022}{Singapore, Singapore}
\acmBooktitle{2022 4th International Conference on Image, Video and Signal Processing (IVSP 2022), March 18--20, 2022, Singapore, Singapore}
\acmPrice{15.00}
\acmDOI{10.1145/3531232.3531235}
\acmISBN{978-1-4503-8741-5/22/03}

\begin{document}

\title{Weakly Supervised Faster-RCNN+FPN to classify animals in camera trap images}

\author{Pierrick Pochelu}
\email{pierrick.pochelu@totalenergies.com}
\affiliation{
  \institution{TotalEnergies SE}
  \city{Pau}
  \country{France}
}

\author{Serge G. Petiton}
\email{serge.petiton@univ-lille.fr}
\affiliation{
  \institution{University of Lille, CNRS, UMR 9189 CRIStAL}
  \city{Lille}
  \country{France}
}

\author{Clara Erard}
\email{clara.erard95@gmail.com}
\affiliation{
  \institution{MODIS}
  \city{Pau}
  \country{France}
}

\author{Philippe Cordier}
\email{philippe.cordier@totalenergies.com}
\affiliation{
  \institution{TotalEnergies SE}
  \city{Paris}
  \country{France}
}

\author{Bruno Conche}
\email{bruno.conche@totalenergies.com}
\affiliation{
  \institution{TotalEnergies SE}
  \city{Pau}
  \country{France}
}

\renewcommand{\shortauthors}{Pochelu et al.}

\ccsdesc[500]{Computing methodologies~Neural networks}
\keywords{Deep Learning, weakly supervised, camera traps, biodiversity monitoring}

\begin{abstract}
Camera traps have revolutionized the animal research of many species that were previously nearly impossible to observe due to their habitat or behavior. They are cameras generally fixed to a tree that take a short sequence of images when triggered. Deep learning has the potential to overcome the workload to automate image classification according to taxon or empty images. However, a standard deep neural network classifier fails because animals often represent a small portion of the high-definition images. That is why we propose a workflow named Weakly Object Detection Faster-RCNN+FPN which suits this challenge. The model is weakly supervised because it requires only the animal taxon label per image but doesn’t require any manual bounding box annotations. First, it automatically performs the weakly-supervised bounding box annotation using the motion from multiple frames. Then, it trains a Faster-RCNN+FPN model using this weak supervision. Experimental results have been obtained with two datasets from a Papua New Guinea and Missouri biodiversity monitoring campaign, then on an easily reproducible testbed.
\end{abstract}
\maketitle

\section{Introduction}

Camera traps due to their non-invasive nature, affordable technology, high mobility, and battery autonomy have revolutionized animal research of many species that were previously nearly impossible to observe due to their habitat or behavior. Several camera traps take hundreds of pictures of animals in specific areas to conduct reliable surveys by identifying them on each. While being one of the main advantages of the technique, this large amount of collected data also proves to be highly time-consuming for ecologists to annotate and count animals.


Taking advantage of the big data era, Deep Learning has had a breakthrough impact on many domains. Since its popularity, deep learning was applied to many colored image datasets often containing several animal classes. Furthermore, deep learning has already shown human-like performance on animal detection. That is why it is a natural choice to help ecologists to leverage the workload.

The Deep Learning image recognition field has evolved in multiple Deep Learning meta-architectures such as:
\begin{itemize}
    \item Faster-RCNN \cite{fasterrcnn:2015} are neural networks able to perform Object Detect tasks i.e., localizing the region of interests and classifying them. However, Faster-RCNN has two major limitations to be applied to camera traps recognition. First, they require high annotation costs including animal localization and animal taxa. Second, they generally fail to detect small objects (here the animals) into a cluttered background. Those neural networks return both animal taxa used to classify and count animals and the localization box which is not useful here and may be ignored.
    \item Classification neural networks require only the taxa annotation, but they also fail to identify objects when the images contain small objects and cluttered backgrounds.
    \item Some methods such as Faster-RCNN with Features Pyramid Network \cite{fpn:2016} (or ``Faster-RCNN+FPN'') are capable to recognize objects of diverse sizes using multiple decisions taken at a different level of abstractions in the neural network architecture. They are specifically suitable to detect small objects in high-definition images but require the same level of annotation as Faster-RCNN.
\end{itemize}


Based on those previous works, we propose Weakly Supervised Faster-RCNN+FPN to combine those two advantages: image-level annotation cost (Weakly Supervised) and the ability to detect small objects (the “FPN” part). This work is presented following this structure: (2) First we introduce the relevant literature regarding the particularity to classify pictures from camera traps: the huge data labeling needs for successful Deep Learning applications and the classification of images of moving objects (here animals). (3) We quickly present both datasets to experiment with our workflow and their inherent challenges for classifying images. (4) The proposed workflow is introduced and discussed. (5) The experimental results of our workflow are compared to baselines and the performances are analyzed.


\section{Related works}

The last 10 years saw the conduct of several projects aiming to classify animals on camera traps \cite{camera:2018}, \cite{swanson:2015}, \cite{willi:2019} but they are often applied to large Savannah animals or livestock where the animal occupy a large portion of the images which is not the case of our datasets.

Previous studies \cite{needle:2019} success to classify images according to a fixed size small objects using a bag of local features \cite{bag:2019}. However, the good performance in practice with varied objects size have not been shown yet.

 

The noise affecting real-world applications make image recognition task more or less challenging for CNNs. In a standard CNN classifier, the lower the signal-to-noise ratio is such cluttered background, small objects to detect (measured with object-to-image), the more labeled data is required \cite{needle:2019}. However, collecting and labeling data is time taking, an ideal system would require only a few thousand classified images to save ecologists’ time. Therefore, an ideal method for this application must be designed to detect small signal signatures in a messy background with a minimum of annotation.

\subsection{Tackle annotation cost}

To reduce the amount of annotation work, we identify three main tracks: extracting knowledge from similarly labeled datasets; using crowd-sourced data labeling software and collaborating with many annotators, or applying weak supervision.

Open dataset \cite{lila} contains terabytes of camera traps images in some areas of the world which can be unlabeled, partially labeled, or contain multiple annotation errors. A popular way to extract useful knowledge from other datasets is referred to as transfer learning, under this umbrella, pre-training weights is a common method  \cite{transfer:2009}.

Another natural method is the usage of crowd-sourced labeling data software \cite{swanson:2015} where many annotators can work together to divide the huge workload out between them. However, only a few ecologists worldwide know to distinguish and annotate some wild animal taxa on a given area of the world.

Another way to tackle annotation effort is to use weak supervision. At the difference of “fully supervised” (or just “supervised”) machine learning models are applied to large datasets with reliable and complete labels. There are 3 kinds of weakly supervised machine learning \cite{weak:2017}: \emph{inaccurate}, i.e., noise affecting labels; \emph{incomplete} (or “semi-supervision” machine learning) i.e., consisting to train an algorithm on both unlabeled and label data samples; \emph{inexact meaning labels} are too coarse-grained in comparison to the task. In Object Detection, the case is said inexact when human experts only label at image-level without qualifying which region of the image is decisive. This is known as Weakly Supervised Object Detection (WSOD).

\subsection{Weakly Supervised Object Detection in images}

Weakly Supervised Object Detection is a method able to provide localization of objects with only class annotation. The first approaches to WSOD formulate this task as a workflow of 3 consecutive steps: \textit{region extraction} to extract candidate regions from the image; \textit{region representation} to compute features representations of regions; \textit{region final decision} such a regressor allows to refine the bounding boxes or a classifier to decide if a bounding box contains an object.

\begin{enumerate}
\item  \textit{Region extraction} to extract candidate regions from the image. The most common region proposals are today the Edge Boxes \cite{edgeboxes:2014}  and the Selective Search methods. \cite{uijlings:2013}
\item \textit{Region representation} to compute features representations of regions such SIFT \cite{lampert:2008}, and HOG \cite{pandey:2011}, and CNN-based \cite{wang:2014}.
\item \textit{Region final decision} to refine the object localization  \cite{pandey:2011} or enable/disable the objectness. More precisely, a regressor allows refining the bounding boxes coordinates or a classifier to decide if a bounding box contains an object \cite{lampert:2008}.
\end{enumerate}

Over the last years, the landscape of weak object detection pushed forward the adoption of efficient end-to-end learning frameworks. Some methods extract the saliency map (i.e., object localization probability map) from a CNN classifier trained with an image-level annotation  \cite{classif1} \cite{classif2} \cite{classif3}. More advanced methods with the same label requirement consist in combining a CNN classifier and, with a saliency map extracted from it, training either another object detection CNN  \cite{wan:2019} or a semantic segmentation CNN \cite{jiang:2013}, \cite{joint:2019}. However, they share in common that the majority of works focus on VOC2007 or VOC2012 datasets where objects are rather centered and occupy a large portion of the image.

Another weak supervision method consists in using a standard strongly supervised model but a labeling rule-based function \cite{snorkel:2017} to automatically build labels with unknown accuracy. For example, JFT-300M dataset was built automatically gathering 375 million images from the web associated with one of the 19 thousand categories depending on web signals. It can be noisy (estimated to ~20\% error) in terms of label confusion and incorrect labels which are not cleaned by humans. This weak supervision machine learning method is popular when it comes to entity extraction where a label function can be found with some relevant hypothesis. If the system is unbiased, we may expect the quantity of automatically labeled data samples may compensate for their low quality.

This label function is domain-specific which is why we propose one to localize animals on camera traps and evaluate its performance. In the `experimental results' section, we compare the complete workflow compared to some relevant baselines. In our raw images, no such heuristic exists to classify and localize our animals in images from camera traps. That is why we propose a new heuristic to localize the animal.

\subsection{Weakly Supervised Object Detection in videos}

Weak supervision is very useful in video recognizing \cite{points:2011} \cite{tranckandsegment:2016} \cite{youreap:2019} containing tens of frames per second. They often rely on motion cues on images and propagate the objectness information over the neighboring pixels (spatial) and neighboring frames (temporal). Generally, the Weakly Supervised Object Detection in videos assumes that relevant objects are similar from frame to frame which is not a relevant assumption in camera traps data where images are taken with a lower time resolution. In a sequence of frames, the animal can be present in only one frame or present on multiple frames but with different positions and different locations.

Match and retrieve regions from the video to an external image dataset have been also used to automatically annotate bounding boxes and their associated class \cite{watchvideo:2016}. This would be difficult to implement in practice in our case because an external training base should contain all animals’ taxa. For instance, only ecologists and a supervised trained neural network can infer efficiently animal taxa.

\subsection{Tiny objects recognition}

Many object detectors including R-FCN, Faster-RCNN are ‘meta-architectures’ and features extractors based on CNN architecture such as VGG or ResNet50. The meta-architecture defines the different modules and how they work together. The feature extractor is a deep learning architecture only made of convolution layers to extract spatial features.

To do object detection, researchers have developed fully integrated meta-architectures like Faster-RCNN \cite{fasterrcnn:2015} which both yield object localization and their classes. The meta-architecture defines the different modules and how they work together. To find an acceptable trade-off between accuracy and computational cost, researchers have proposed SSD or YOLO which are especially popular to recognize objects in videos, that is to say, millions of low-resolution images or usable for real-time applications. Our need is to recognize a few thousand highly detailed images containing often challenging small objects. That is why we give priority to accuracy over inference speed in this work, Faster-RCNN is a known robust meta-architecture containing three specialized trainable modules trained end-to-end. The CNN extractor is a standard deep learning architecture without any classification layer which extracts spatial features shared with the Region Proposal Network (RPN). The RPN is trained to localize Region of Interests (ROIs) such as the position of the animal in the picture and it sends them to the “classifier and regressor module”. This last module classifies each ROI and is trained to classify the localized animal and also to refine the ROIs coordinates.

\begin{figure}[h]
  \centering
  \includegraphics[width=0.9\linewidth]{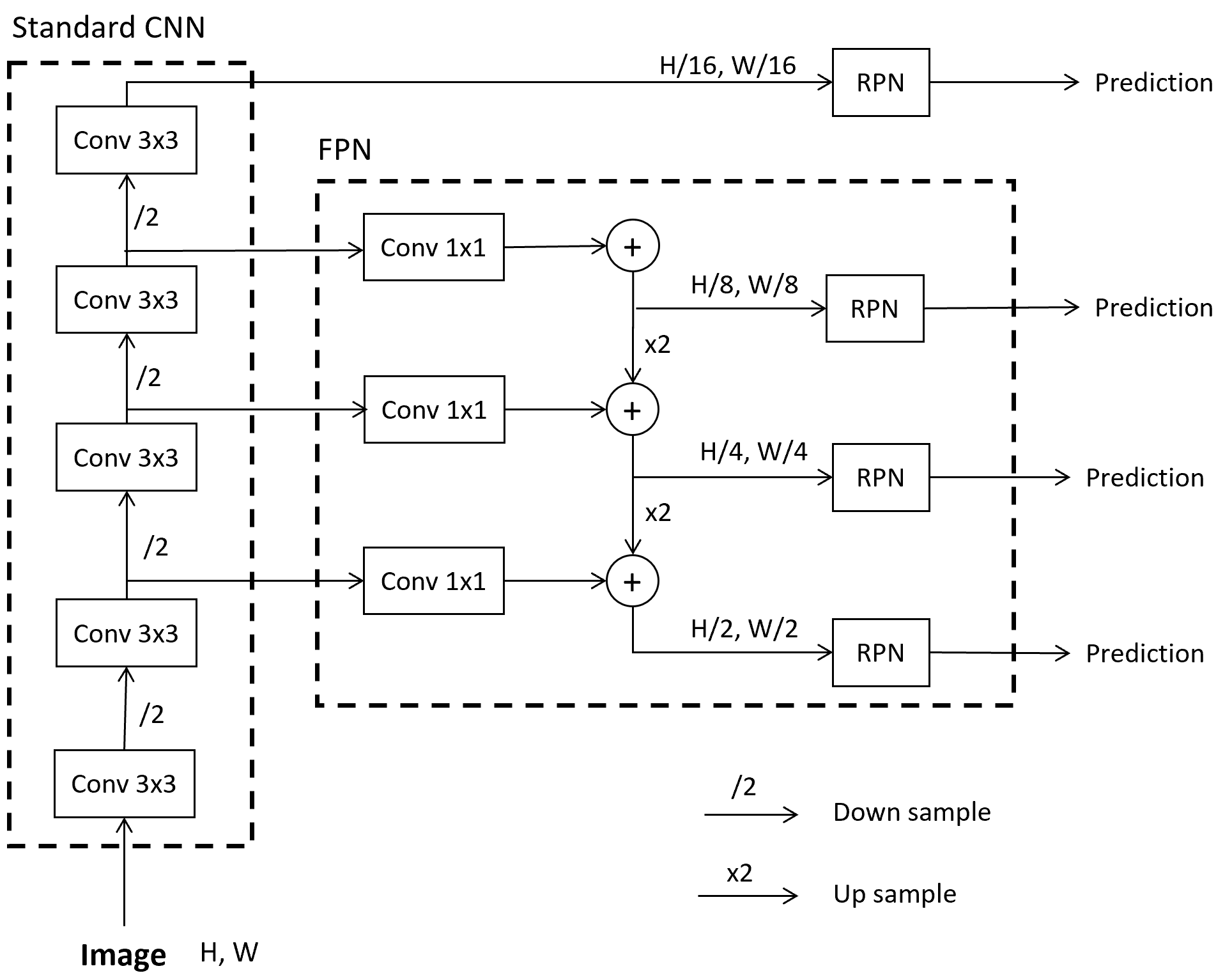}
  \caption{The Faster-RCNN+FPN (Features Pyramid Network) meta-architecture. ``+'' symbols indicate the addition between two information flows. }
  \label{fig:fpn}
\end{figure}

In theory, standard CNN is adapted to detect different levels of abstraction in the input image. In practice, it fails to preserve tiny signals in the input image of the greatest importance on the final prediction. FPN \cite{fpn:2016} presented in \ref{fig:fpn} is trained to extract and make decisions on different levels of abstraction and resolution (the features pyramid). Faster-RCNN with FPN shows a significant improvement in several applications compared to basic Faster-RCNN models.


\subsection{Summary}

In summary, this paper is built on previous research on weak supervision methods and object detection to recognize small objects. The goal is to save the annotation effort on camera traps and to benefit from the state-of-the-art supervised object detection method to spot tiny objects (animals). While most recent initiatives are focusing on how to apply WSOD to localize objects with an end-to-end CNN classifier with some reasonable size objects (VOC2007, VOC2012) or video with multiple similar frames, we rather attempt to build a labeling function applied to small moving objects on short sequences of images

\section{The method}

Our model is the fruit of those past previous works combined with some domain specific functions.  It uses a meta-architecture Faster-RCNN \cite{fasterrcnn:2015} with Features Pyramid Network \cite{fpn:2016} and ResNet50 features extractor \cite{resnet:2015}. It is pre-trained \cite{transfer:2009} with the ImageNet dataset containing 1,000 classes of which 398 are animal classes. Pre-training allows to reduce by about 5 times the convergence speed, but we observe it does not make the training converges significantly higher. The goal of this work is to classify images to count animal occurrences. To measure the suitability of our method we use the percentage of well-classified images compared to its class, the class is the animal in taxa or the reject class “empty” when no animal is present in the photo. Indeed, the exact position of the bounding boxes of Faster-RCNN does not count in our target metric.

When several instances of the same species are in the same image, the taxon must be correctly identified but we do not count them. Due to the scarcity of inter-species social behavior, we witnessed only once different animal taxa on the same photograph in the Papua New Guinea dataset. In this case, in the evaluation phase, we count 2 possible correct predictions, and in the training phase, we only label the biggest animal (pixel squared).

\subsection{The workflow usage}

Before applying deep learning, the ecologists manually check all images and they input a mapping file that maps the name of the photograph with a taxon number or ‘0’ if there is no animal in the photo. Now, using our proposed system, the ecologist labels a few thousand images, then he/she runs the training phase before running the deep learning inference model on the remaining images.

Because no deep learning model is perfect, to save ecologists’ time and reduce errors, all predictions of the systems are sorted in ascending order to their Softmax posterior probability. Therefore, the first images have more chance to contain challenging images to recognize, like taxa absent from the training dataset. After that, the ecologist goes through the predictions and associated images and corrects the mapping file when needed.

\subsection{Weakly Supervised Object Detection proposed workflow}

The main criticism against deep learning is the long labeling time needed for the algorithms to be accurate. The advantage of the proposed classification Deep Learning method is that only one piece of information is needed by an animal in a picture - its taxon - which makes it faster and easier to label training images.

\begin{figure*}[h!]
  \centering
  \includegraphics[width=0.75\linewidth, trim=0 0 5cm 0,clip]{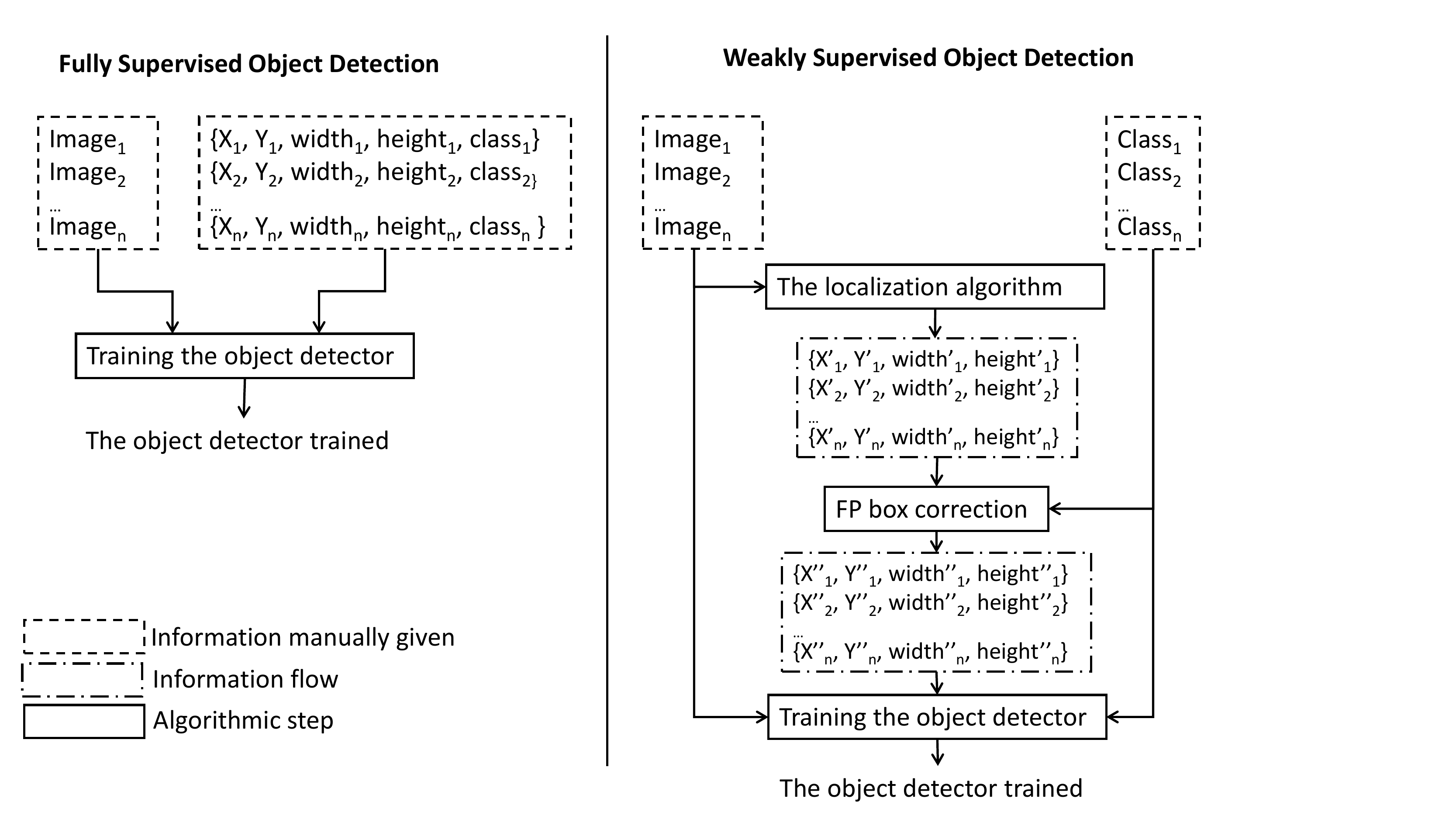}
  \caption{Comparison between the usual supervised object detection and the proposed Weakly Supervised Object Detection workflow.}
  \label{fig:sub2}
\end{figure*}

Our weakly supervised Faster-RCNN+FPN method is winning on both sides: it can accurately localize and classify a small signal in a cluttered background thanks to the FPN module while requiring classification labeling effort (taxa) and not the usual object detection effort (localization annotations and taxa).  The overall proposed method is shown in figure~\ref{fig:sub2} and described in the following paragraphs.

\textbf{The localization algorithm.}  The presence of burst mode on modern camera traps makes it possible to capture the same animal in a short sequence of frames in a few seconds when the camera is triggered. Therefore, the animal motion in those frames is possible to distinguish from the background. A motion-based localization allows the ecologists to handle the workload induced by automatically computing bounding boxes of the animals and it enables to feed the object detection neural network.

\begin{figure*}[h!]
\centering
    \includegraphics[width=0.8\linewidth]{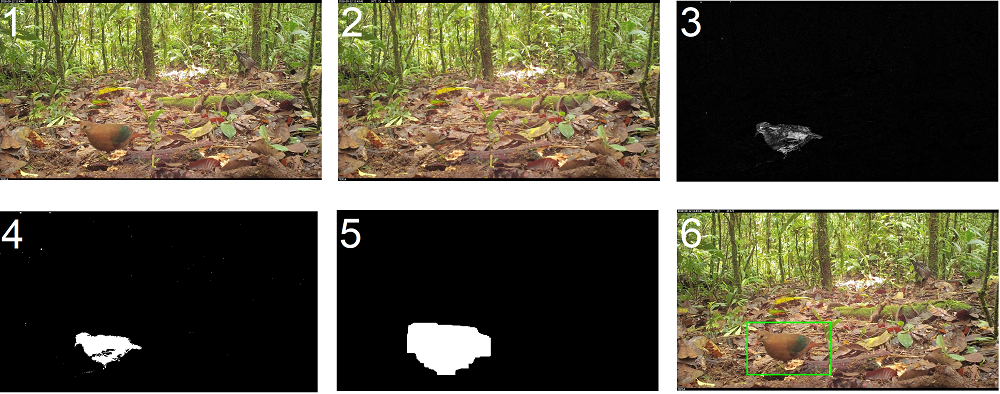}
    \includegraphics[width=0.48\linewidth]{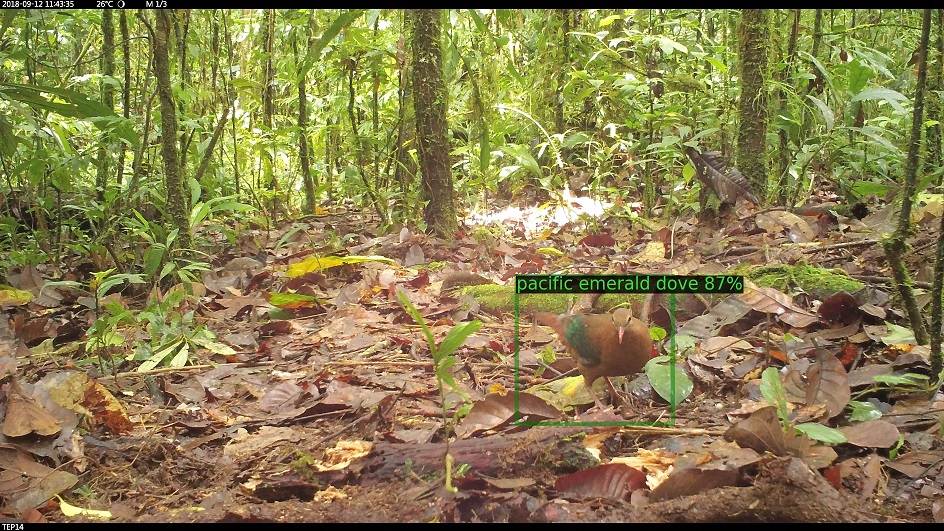}
  \caption{Our localization algorithm proceeds running step-by-step. The pacific emerald dove is detected despite its camouflage. After training on images like the sixth one, Faster-RCNN+FPN can be accurate on similar images.}
  \label{fig:lbscca}
\end{figure*}

Our localization algorithm proceeds following those 6 steps and they are illustrated in figure \ref{fig:lbscca}: 
\begin{enumerate}
    \item Input a short sequence of images $I_{1}, I_{2}, ..., I_{n}$.
    \item It computes a background $B$ computed with median filtering of all the $n$ images of the burst.
    \item Motion map $M_{1}, M_{2}, ..., M_{n}$ are euclidean distances between each pixel into $I_{1}, I_{2}, ..., I_{n}$ and each associated pixel in $B$.
    \item A binary threshold $t$ is applied on all motion maps $M$ with t=12\%. Those new maps named $T$ may contain salt and pepper noise.
    \item A morphological opening operation is applied on those previous binary maps $T$. First an erosion operation with kernel 3x3 allows us to erase noisy connected components. Then, a dilation operation with kernel 151x151 ensures all animal parts are connected. The denoised motion maps are named $D$.
    \item The bounding boxes of the largest connected components (if any) are computed from $D$ and returned.
\end{enumerate}

The accuracy of the localization algorithm is 77.6\% on the Papua New Guinea dataset test dataset, with 14.7\% of false-positive error (motion from raindrop, flying bugs, or wind in the vegetation) and 7.7\% of false-negative error (The animal motion is missed).

\textbf{The FP box correction algorithm.} All the FP errors are automatically corrected based on a simple if-then-else code structure. It compares the presence of a detected box and the presence of an animal in the class label to avoid FP errors. Thus, Faster-RCNN+FPN is trained on 92.3\% of correct labels (77.6\%+14.7\%).

The time performance of the workflow depends mostly on the hardware, the resolution of images, and the implementation. Our 3,776x2,124 resolution images are predicted with a throughput of 1.78 images/sec on an NVIDIA Tesla V100 GPU inside an NVIDIA DGX1 computer. Regarding previous studies on mammalian species in the African Savannah \cite{swanson:2015}, it took over 28,000 registered citizen scientists between 2 and 3 months to classify 3.2 million images. Our code would take 21 days with one single modern GPU with equivalent resolution. An actual production code would be faster than the current inference throughput due to software optimizations.

\section{The two datasets and goal}

This section provides detailed descriptions of both datasets used as a benchmark and their inherent challenges. Reading this section is not mandatory to understand the nature of our work.

\subsection{Papua New Guinea biodiversity}

Those images are collected from a monitoring campaign that took place in Papua New Guinea. 8 motion-triggered wildlife camera traps have been used for 15 months with the same settings. They have been operated tied to a tree from 43 to 101 days and took between 474 and 1,272 photographs. When triggered, they shot a burst of three photographs spaced by 1 second, increasing the chances to have at least one good image of the animal. As a majority of mammals are nocturnal, camera traps are also equipped with a flash mode. To decrease potential errors made by our localization algorithm, tuning the settings of cameras is an axis of improvement including the number of shots per sequence and the delay between shots in one sequence.

25 taxa have been captured (dorcopsis luctuosa, bandicoot, scrubfowl, emerald dove...) into 5400 images with 8 cameras. Each camera is fixed to a tree and captures similar background when it is triggered, even if the background may slowly evolve due to leaves falling, weather, flowers blooming ... We want our system to be trained on some camera traps and may be able to generalize on new ones. 6 cameras contain 4,320 labeled pictures, 95\% of sequences serve as training and the remaining ones are the validation dataset. The testing dataset contains 1,080 pictures taken by the 2 remaining cameras. All pictures are colored with a resolution of 3,776x2,124 pixels. Each image is associated with one label, those labels are edited in a file which maps the identifier number of the picture and the identifier number of the class. If a picture contains an animal, the class identifier is set to the animal taxa starting at ‘1’. Otherwise, the class identifier is set to ‘0’.

\subsection{Missouri biodiversity}

This open-access dataset comes from the Missouri campaign \cite{missouri:2021} \cite{zhang:2016}. It contains approximately 24,673 camera trap images. There are about 1277 fully labeled images with bounding boxes and classes (squirrel, European hare, red deer, red fox, ...). Among them, a statistic we made on 100 randomly drawn images shows 10\% false-positive errors (bounding box and animal taxon whereas we do not see animal into), no false-negative errors, and a majority of badly framed bounding box annotations. The tiny portion of the labeled dataset and the fact that annotations contain errors make the task particularly challenging. We checked and corrected 1000 images to create the test dataset.

\subsection{Computer vision challenges}

Wildlife camera trap automatic recognition present some challenging situation for deep learning methods. First, the animals are small in a cluttered background making it difficult to recognize them. We measure on the Papua New Guinea dataset that the surface of animals can occupy from the entire image to $<0.2\%$ of pixels of the photo depending on its actual size and its proximity to the camera, with a median of 4\%. Additionally, both datasets contain unbalanced classes. 

In addition, we summarize the difficulty to recognize an animal in five challenging situations: when the animal is hidden, when the animal is badly framed, when the animal appear blurred (e.g., motion blur when it is running or jumping), hidden behind foliage, unusually small ($<0.2\%$ of pixels of the image), when the animal is unknown to the training dataset. They are illustrated in supplementary material section~\ref{app:chal}.

\section{Experimental results}
\label{sec:results}

In this section, we compare different workflows in terms of accuracy. Then we make an in-depth analysis of its errors on the Papua New Guinea dataset. Finally, we assess different classification methods on a testbed by varying the object-to-image rate (O2I).  

\subsection{Comparison}

The accuracy comparison of different workflows is shown in table~\ref{tab:methods}. Those workflows, their settings, and their characteristics are discussed in the following text.

\textbf{CNN Classifier}. It is a trained classification deep learning model ResNet50 \cite{resnet:2015}. It takes as an input the overall photograph and classifies among 26 classes, 25 animal taxa, or the empty class. We tested also VGG16, InceptionV3, EfficientNet-B4 with similar settings: 50 epochs and batch size of 32. ResNet50 is more accurate than VGG16 and performs the same as InceptionV3. The neural network is trained with SGD and the learning rate starts at $10^{-3}$ and is divided by 10 after the $20^{th}$ and the $40^{th}$ epochs. Despite we spend time assessing many neural network architecture and optimizer settings, we have not obtained satisfactory results. It takes only 1 hour to be trained, thus this workflow and RP+Classifier are the fastest to converge.

\textbf{RP+Classifier.} (Region proposals+Classifier) This method uses also a ResNet50 classifier with the same training set as the CNN Classifier workflow. The difference is that it is trained on the region of interests yielded by the localization algorithm presented in  section~\ref{fig:lbscca} and the overall image such as Faster-RCNN and classifier models. An empty class is added because the localization algorithm provides frequent (14.7\%) false-positive error motions (again, wind, flying bugs, ...). We explain the performance gained compared to the standard classifier by the fact that the localization algorithm crops most of the cluttered background, thus the classifier part is trained and predicts with a better focus on the moving object.

\textbf{Auto-Keras.} (version 1.0.12) It is an AutoML strategy named Auto-Keras Image Classifier \cite{autokeras:2019}. It is a Bayesian optimization algorithm searching the best neural network architecture using validation metrics and calibrating the weights of the candidate on the training dataset. We observe a small improvement compared to the ResNet50 classifier after tuning 100 models trained a maximum of 20 epochs but the training computing cost is about multiplied by 100 (4 days on 1 NVIDIA Tesla GPU). It shows that the famous ResNet50 neural architecture is relevant in our datasets and spending days searching for a specific neural architecture may only improve the results a little bit.

\textbf{Weakly Supervised Faster-RCNN.} We use ResNet50 inside Faster-RCNN and Faster-RCNN+FPN meta-architecture. The neural network converges after 200 epochs during 4 hours on Tesla GPU with a batch size of 32. The model is trained with SGD and the learning rate starts at $10^{-3}$, then divided by 10 after the $100^{th}$, $170^{th}$ and $190^{th}$ epochs. In the case of the Papua New Guinea dataset, we compare our workflow with and without manually correcting the bounding box annotations. And in the case of the Missouri dataset, we compare with our workflow and the original bounding boxes downloaded with the Missouri dataset.

\begin{table*}[]
\setlength\tabcolsep{1.5pt} 
\centering
\caption{Summary of the different deep learning methods tested and their performance on two datasets Missouri (MIS) and Papua New Guinea (PNG). Regarding the method, the localization refers either to class if an animal is present in the picture (in the case of classifiers) or to localize its bounding boxes (in the case of Faster-RCNN models). (1) our localization algorithm with handcraft corrections, (2) our localization algorithm motion-based (3)  localization algorithm by \cite{zhang:2016}}
\begin{tabular}{cllllll}
~                   & ~          & ~                 & \multicolumn{2}{c}{Presence err.} & \multicolumn{2}{c}{Presence OK} \\
Data                   & Annotation          & Method                 & FN & FP & Taxa error & Acc. \\
\hline
\multirow{6}{*}{PNG} & Supervis. classif. & CNN classifier         & 31.4                                                  & 6.7                                                   & 34.3                                                             & 27.6     \\
                     & Supervis. classif. & Auto-Keras             & 28.7                                                  & 5.1                                                   & 37.6                                                             & 28.6     \\
                     & Weak Obj. Det.(2)      & RP+Classifier      & 7.7                                                   & 11                                                    & 10.3                                                             & 71       \\
                     & Weak Obj. Det.(2)      & F-RCNN            & 27.7                                                  & 18.1                                                  & 12                                                               & 42.2     \\
                     & \textbf{Weak Obj. Det.(2)}      & \textbf{F-RCNN+FPN(ours)} & \textbf{2.8}                                                   & \textbf{1.7}                                                   & \textbf{13.8}                                                             & \textbf{81.7}     \\
                     & Supervised O.D.(1)   & F-RCNN+FPN    & 1.8                                                   & 1.6                                                   & 10.3                                                             & 86.3     \\
\hline
\multirow{6}{*}{MIS} & Supervis. classif. & CNN classifier         & 24.2                                                  & 12.5                                                  & 30.9                                                             & 32.4     \\
                     & Supervis. classif. & Auto-Keras             & 20.2                                                  & 11.7                                                  & 34.4                                                             & 33.7     \\
                     & Weak Obj. Det.(2)      & RP+Classifier      & 19.9                                                  & 8.5                                                   & 22.5                                                             & 49.1     \\
                     & Weak Obj. Det.(2)      & F-RCNN            & 26.4                                                  & 10.4                                                  & 22.6                                                             & 40.6     \\
                     & \textbf{Weak Obj. Det.(2)}    & \textbf{F-RCNN+FPN(ours)} & \textbf{15.9}                                                  & \textbf{4.2}                                                   & \textbf{20.5}                                                             & \textbf{59.4}     \\
                     & Weak Obj. Det.(3)    & F-RCNN+FPN     & 18.1                                                  & 4.6                                                   & 27.3                                                             & 50 
\end{tabular}

\label{tab:methods}
\end{table*}

The downloaded dataset annotation of the Missouri dataset ((3) in the figure \ref{tab:methods}) \cite{zhang:2016} contains inaccurate bounding box annotations. It suffers from about 10\% images which contain wrong animal labels or boxes without a visible animal in them. Our proposed workflow can cancel a bounding box when the class of the image is ``empty'' but it cannot handle when a wrong animal class is given and thus the biggest move (e.g., the wind on the grass) is used as a bounding box associated to a false taxon. We run our Localization based on the motion on this same dataset and provide better annotations ((2) in the figure \ref{tab:methods}).

RP+Classifier works better than classifier thanks to its ability to focus on the region of interest but is still inferior to Faster-RCNN+FPN. RP+Classifier loses contextual information: the relative size of the region of interest is lost because it is resized to the fixed size 256x256 to feed the classifier. And more, the surrounding is also lost it may be a major issue when the localization poorly frames the animal, thus some animal parts are cropped and not given to the neural network. 

Finally, we observe on the Papua New Guinea dataset that fully supervised deep learning performs +4.6\% on the test compared to the weakly supervised counterpart (again, 7.7\% of the training dataset is affected by false-negative error boxes).

\subsection{In-depth analysis of challenges}
\label{sec:analchal}

Table~\ref {tab:res4} shows that our method displays an overall accuracy of 81.7\%. In practice, when an error is made, other images in the sequence often contain at least once the right animal identification.

\begin{table}[h]
\small
\centering
\setlength\tabcolsep{2pt} 
\caption{Distribution of the different challenges and errors on Papua New Guinea dataset. The ``tiny'' tag is formally defined by O2I$<0.2\%$ which are unusually small regions.}
\begin{tabularx}{\linewidth}{llllllllll}
\cline{3-8}
                      & \multicolumn{1}{l|}{}                                                                                & \multicolumn{6}{c|}{Data test photograph characteristic}                                                                                                                                                                                                                                      &          &  \\ \cline{3-8}
                      & \multicolumn{1}{l|}{}                                                                                & \multicolumn{1}{l|}{\begin{tabular}[c]{@{}l@{}}good \\ image\end{tabular}} & \multicolumn{1}{l|}{\begin{tabular}[c]{@{}l@{}}bad \\ framed\end{tabular}} & \multicolumn{1}{l|}{blur}  & \multicolumn{1}{l|}{hidden} & \multicolumn{1}{l|}{tiny}  & \multicolumn{1}{l|}{unknown} & sum      &  \\ \cline{2-8}
\multicolumn{1}{l|}{} & \multicolumn{1}{l|}{correct}                                                                         & \multicolumn{1}{l|}{62.5\%}                                                & \multicolumn{1}{l|}{9.3\%}                                                 & \multicolumn{1}{l|}{5.3\%} & \multicolumn{1}{l|}{3.8\%}  & \multicolumn{1}{l|}{0.7\%} & \multicolumn{1}{l|}{0.0\%}   & \textbf{81.7\%}   &  \\ \cline{2-8}
\multicolumn{1}{l|}{} & \multicolumn{1}{l|}{\begin{tabular}[c]{@{}l@{}}good\\ localization,\\ badly classified\end{tabular}} & \multicolumn{1}{l|}{2.2\%}                                                 & \multicolumn{1}{l|}{3.0\%}                                                 & \multicolumn{1}{l|}{1.2\%} & \multicolumn{1}{l|}{0.2\%}  & \multicolumn{1}{l|}{0.5\%} & \multicolumn{1}{l|}{6.8\%}   & 13.8\%   &  \\ \cline{2-8}
\multicolumn{1}{l|}{} & \multicolumn{1}{l|}{\begin{tabular}[c]{@{}l@{}}False negative\\ localization\end{tabular}}                     & \multicolumn{1}{l|}{0.5\%}                                                 & \multicolumn{1}{l|}{0.3\%}                                                 & \multicolumn{1}{l|}{0.2\%} & \multicolumn{1}{l|}{0.8\%}  & \multicolumn{1}{l|}{0.8\%} & \multicolumn{1}{l|}{0.2\%}   & 2.8\%    &  \\ \cline{2-8}
\multicolumn{1}{l|}{} & \multicolumn{1}{l|}{\begin{tabular}[c]{@{}l@{}}False positive\\ localization\end{tabular}}                     & \multicolumn{1}{l|}{1.7\%}                                                 & \multicolumn{1}{l|}{-}                                                 & \multicolumn{1}{l|}{-} & \multicolumn{1}{l|}{-}  & \multicolumn{1}{l|}{-} & \multicolumn{1}{l|}{-}   & 1.7\%    &  \\ \cline{2-8}
                      & sum                                                                                                  & 66.8\%                                                                     & 12,7\%                                                                     & 6.7\%                      & 4.8\%                       & 2.0\%                      & 7.0\%                        & 100,00\% & 
\end{tabularx}
\label{tab:res4}
\end{table}

We do our best to break down all the test datasets into those types of challenging images but choosing for each image is sometimes subject to interpretation. However, we show that the majority of errors are caused by those challenging images and 7\% of errors are unavoidable due to the discovery of new species. We also show that 1/3 of images are challenging.

\subsection{Tiny object recognition}

We observe in the previous section that on two datasets Faster-RCNN+FPN was superior to other approaches. To better understand this accuracy, we use the nMNIST testbed \cite{needle:2019}. It is an easily reproducible testbed that consists to classify if the image contains or not the digit `3' in a randomly generated cloud of digits. The experiment is repeated with four different O2I (object-to-image): \{19.1\%, 4.8\%, 1.2\%, 0.3\%\} and the corresponding number of digits in each image generated is \{3, 6, 26, 101\}. For all values of O2I ratio, 11276, 1972, 4040 of training, validation, and testing images, out of which 50\% are negative and 50\% are positive images.

We report results of the workflow previously proposed \cite{needle:2019} and meta-architectures. We observe roughly similar results when changing the architecture (such replace ResNet50 with EfficientNet-B4) so we decided do not to show them to keep the figure readable.

\begin{figure}[h!]
\centering
    \includegraphics[width=0.9\linewidth]{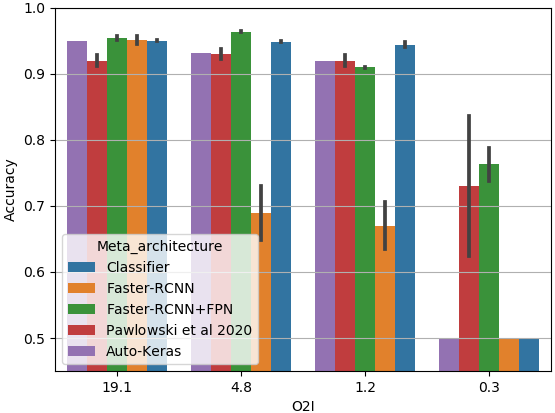}
  \caption{We apply 5 workflows on the nMNIST testbed. They all be ran at least 3 times and the standard deviation plotted, at the exception of Auto-Keras which is run only once because of prohibitive run time.}
  \label{fig:testbed}
\end{figure}

We observe that when O2I=19.1\%, all workflows perform around 95\% but [Pawlowski et al 2020] which is around 92\%. When O2I=0.3\%, Faster-RCNN+FPN outperforms all other workflows with a 76.83\% accuracy, and [Pawlowski et al 2020] shows very unstable results with different run times. More surprisingly, when decreasing O2I, Faster-RCNN without FPN accuracy drops faster than all the other workflows. However, on our biodiversity datasets, Faster-RCNN performs better than classification neural networks.

Those experiments confirm that Faster-RCNN+FPN is the more stable to recognize needle-in-haystacks and performs also very well when the object appears bigger. In camera trap applications, due to the accuracy of Faster-RCNN+FPN, we conclude it deserves to use an additional localization algorithm to compute bounding box annotations to feed it.

\section{Future works}

An important line of research consists in evaluating uncertainty estimates allowing to focus ecologists' attention on the most uncertain predictions (badly framed, blurred, unknown species, ...). Ensemble Deep Learning has already been shown to be not only useful to boost the accuracy but also to produce qualitative uncertainty estimates \cite{uncertainty:2017}. However, those methods multiply the computing cost over one single neural network. It seems only experiments and throughout analysis may study the balance between costs and benefits of ensembling.

Results obtained in this paper show us that Faster-RCNN+FPN suits the ability to recognize small objects into a rather fixed cluttered background. One disadvantage of our workflow is the need for a localization algorithm, even if it shows robust results on two datasets, it is a non-trainable part that cannot properly work if the background contains background motions (such as a city) or if the quality of the overall image change between shots (blur, contrast, ...). Our future works will compare the presented workflow with an end-to-end neural network where we input all images in a sequence to a model at once to allow the model to classify each frame by also using the temporal context. It may be possibly able to focus automatically on the small moving parts.

\section{Conclusion}

Deep learning has led to progress in image recognition and has been tested extensively to identify animals and yet its application to biodiversity monitoring is still in its infancy. The limitation of the application to a real need comes from the need to have a dataset previously labeled for a given region. The successful application of our weakly-supervised Faster-RCNN with Feature Pyramid Network addresses the need to recognize small objects with a few thousand labeled images. Compared to the fully supervised counterpart we divide by factor 5 the amount of information (the taxon and 4 box coordinates) and the accuracy drops by less than 5\%. Now the ecologists can check if the predictions are correct in priority where the model gives a high uncertainty estimate. This allows focusing on the most challenging images or animals absent from the training dataset. It is by this means that in our campaign we discover the Palm Cockatoo taxon which was absent from our initial training dataset. This method could boost camera traps adoption by tackling the inherent challenges. Additionally, we also hope it will shed light on the benefits of weakly supervised deep learning methods for all disciplinary communities.




\begin{acks}
We would like to thank TotalEnergies SE and its subsidiaries for allowing us to share this material and make available the needed resources. We also thank Aurelien Puiseux for labeling the Papua New Guinea dataset.
\end{acks}

\bibliographystyle{ACM-Reference-Format}
\bibliography{bib_biodiv,bib_cvpr,bib_dl}


\clearpage

\section*{Challenging image types}
\label{app:chal}

\setcounter{figure}{0}                       
\renewcommand\thefigure{A.\arabic{figure}}   

\begin{figure*}[h]
\centering
\begin{minipage}{0.49\linewidth}
\centering
\includegraphics[width=0.99\linewidth]{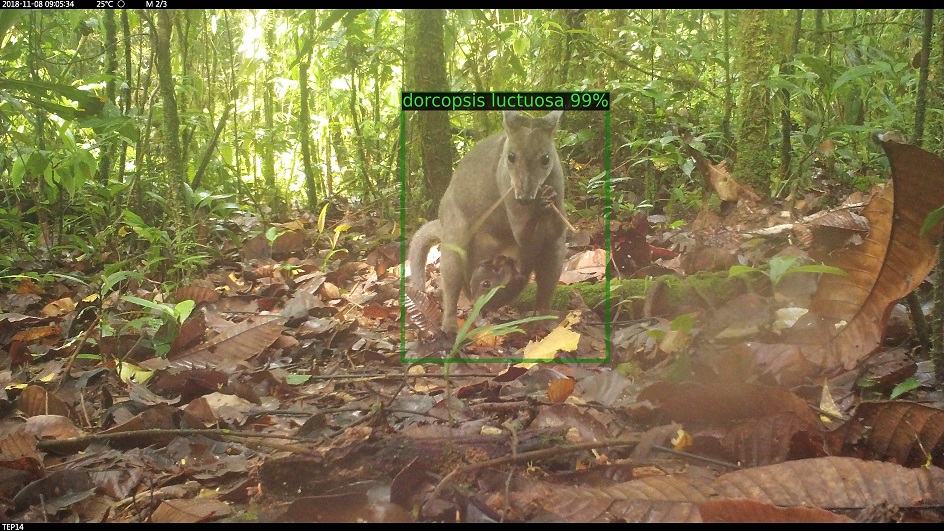}
\end{minipage}\hfill
\begin{minipage}{0.49\linewidth}   
\centering 
\includegraphics[width=0.99\linewidth]{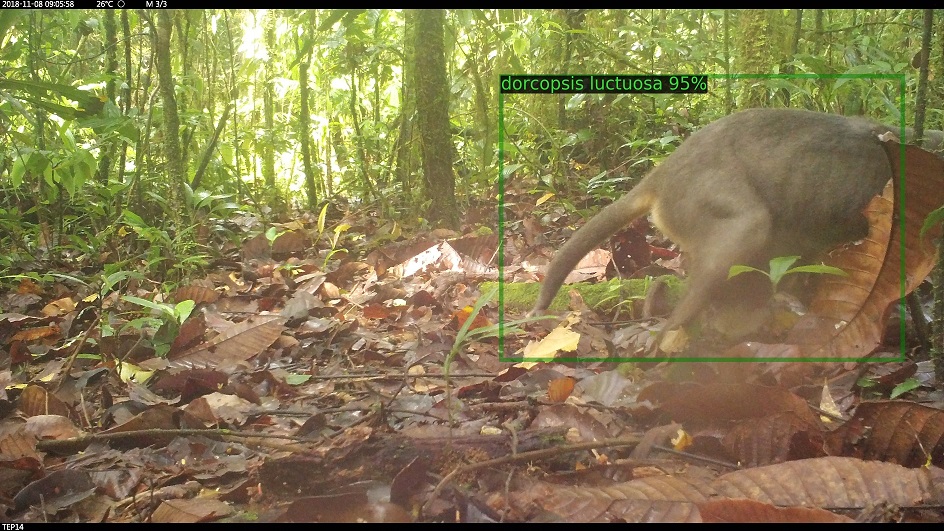}
\end{minipage}
\caption{\textbf{Badly framed mammals} (right) are difficult to distinguish because many mammals have brown or grey fur in this area. To compare, a good image of ``dorcopsis luctuosa'' taxon is shown (left). The two predictions are correct.}
 \label{fig:fur}
\end{figure*}

\begin{figure*}[h]
\centering
\begin{minipage}{0.49\linewidth}
\centering
\includegraphics[width=0.99\linewidth]{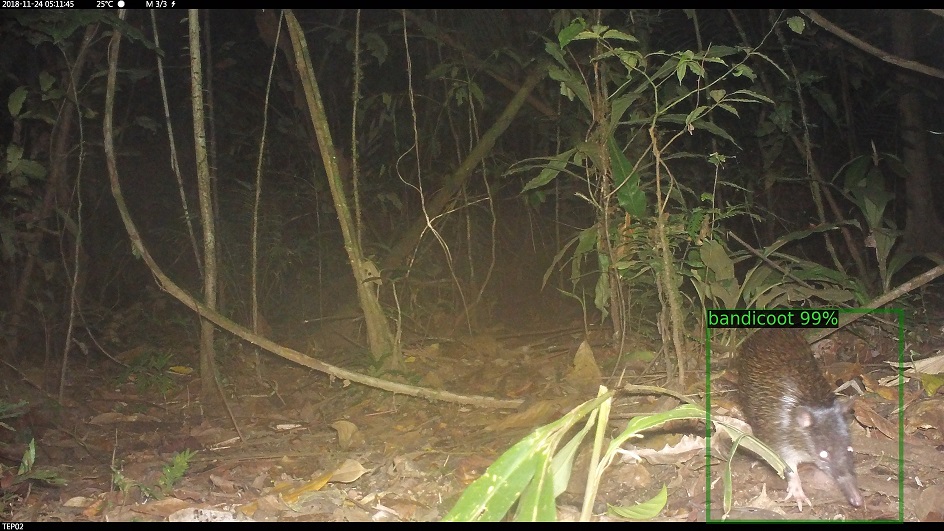}
\end{minipage}\hfill
\begin{minipage}{0.49\linewidth}   
\centering 
\includegraphics[width=0.99\linewidth]{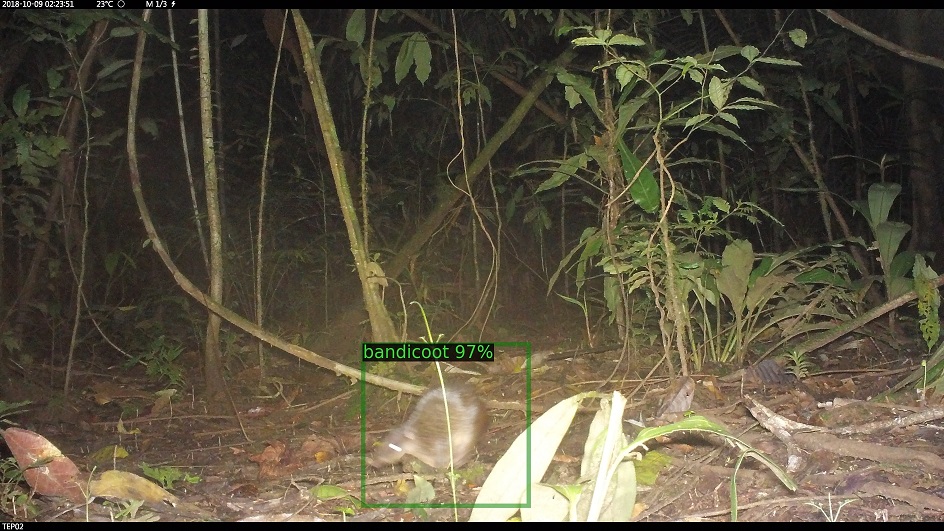}
\end{minipage}
\caption{\textbf{Motion blur} effect (right). To compare, a good image of ``bandicoot'' taxon is shown (left). The two predictions are correct.}
 \label{fig:fast}
\end{figure*}

\begin{figure*}[h]
\centering
\begin{minipage}{0.49\linewidth}
\centering
\includegraphics[width=0.99\linewidth]{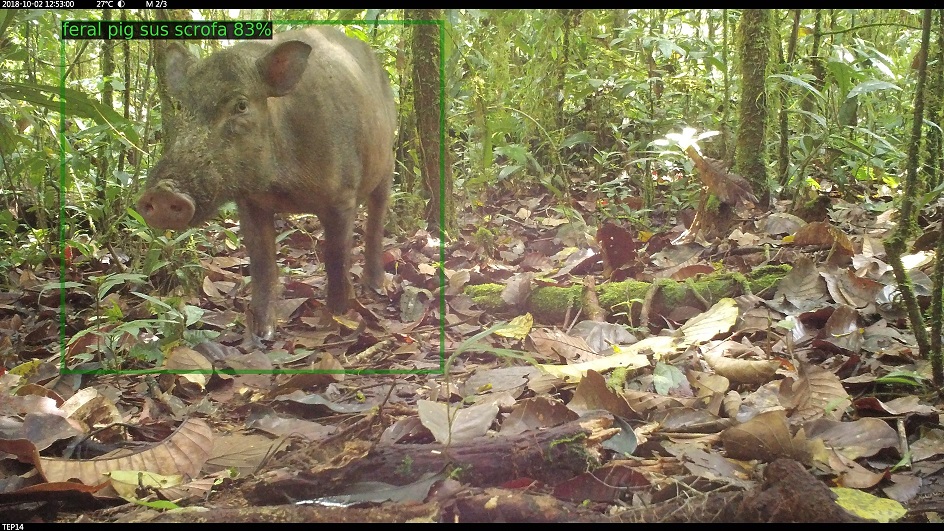}
\end{minipage}\hfill
\begin{minipage}{0.49\linewidth}   
\centering 
\includegraphics[width=0.99\linewidth]{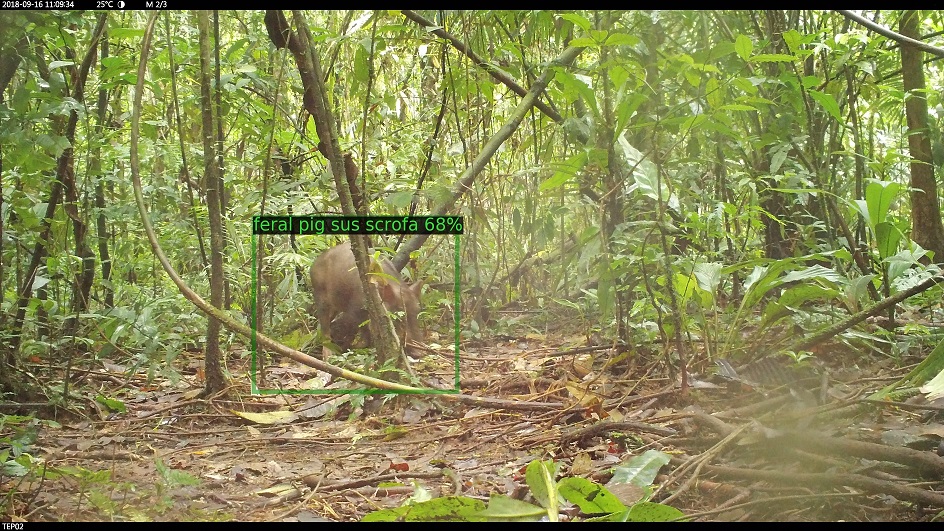}
\end{minipage}
    \caption{\textbf{Occlusions} of leaves and branches (right) can hide important parts of an animal that are necessary to identify it. To compare, a good image of ``feral pig sus scrofa'' taxon is shown (left). The two predictions are correct.}
  \label{fig:hide}
\end{figure*}

\clearpage

\begin{figure*}[h]
\centering
\begin{minipage}{0.49\linewidth}
\centering
\includegraphics[width=\linewidth]{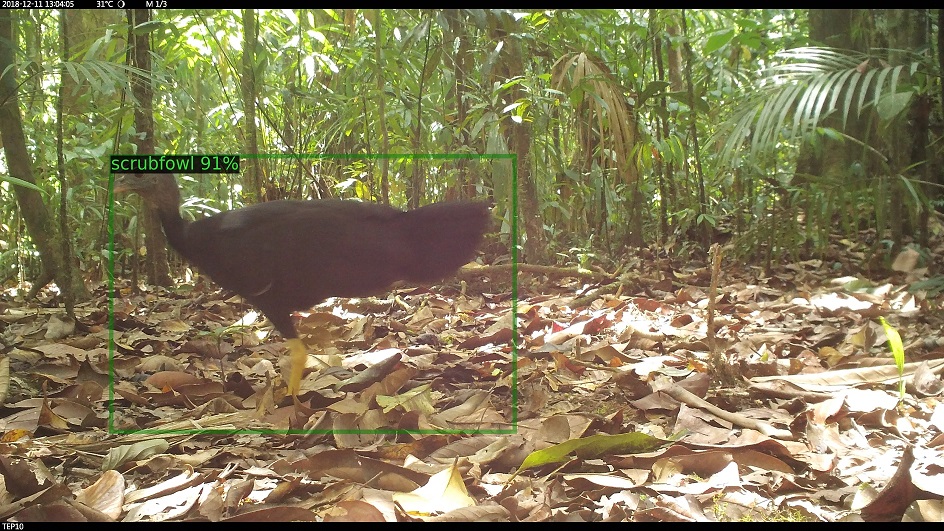}
\end{minipage}\hfill
\begin{minipage}{0.49\linewidth}   
\centering 
\includegraphics[width=\linewidth]{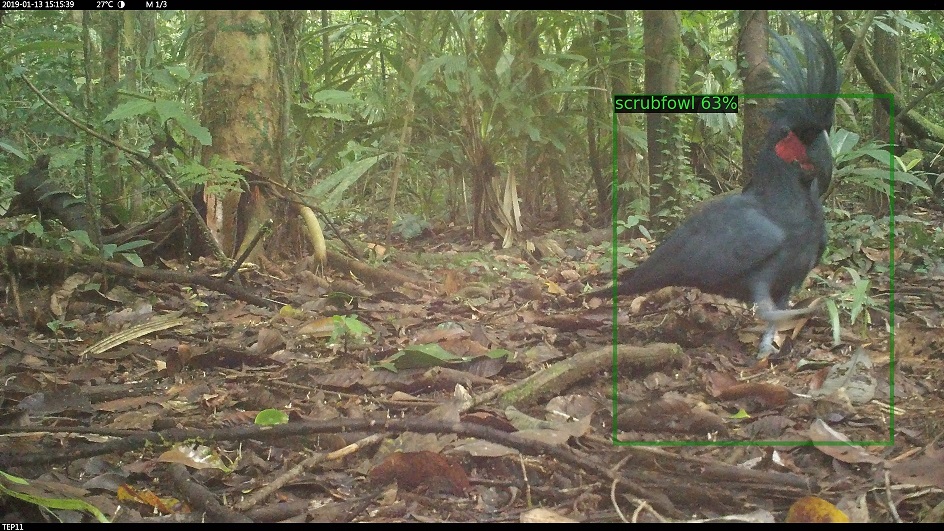}
\end{minipage}
    \caption{\textbf{In absence} of a taxon in our training dataset, the classifier will classify them among known classes leading to an avoidable error. The ``palm cockatoo'' taxon (right) was not photographed during our previous campaigns.}
  \label{fig:rare}
\end{figure*}

\vspace{1cm}

\begin{figure*}[h]
\centering
\includegraphics[width=0.8\linewidth]{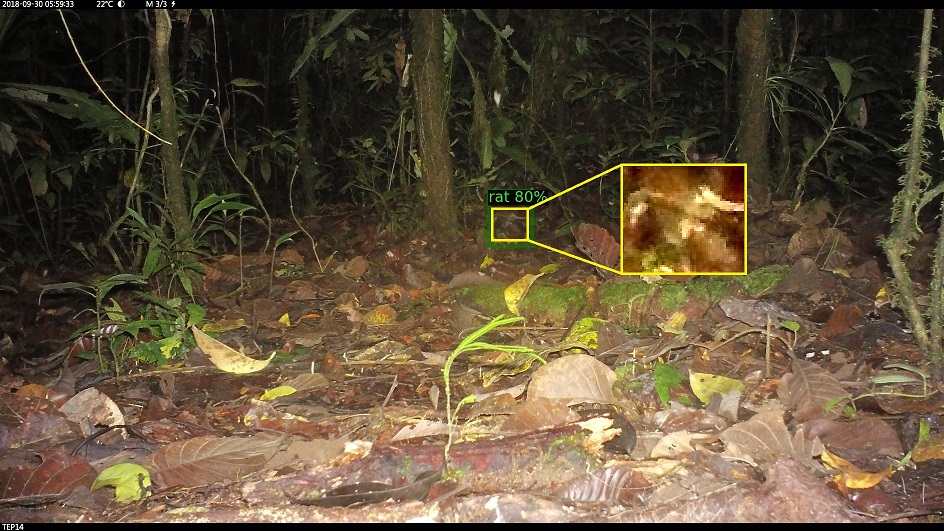}
\caption{\textbf{Tiny} pixel representation of a ``rat'' taxon. The rat seen from the back is running and jumping in the opposite direction of the camera.  The right image rat representation takes only 122x110 pixels. It makes 0.17\% of the surface of the overall image because not only is it a small mammal but it is also far from the camera. Contrast enhancement and magnification allow us to see it: its head is in in the right bottom corner, its tail above its head and back paws do not touch the ground. We can enumerate three reasons for this high level of accuracy: Imagenet pre-training already contains rat images, we collect also many rat images in our training dataset, finally, Faster+FPN is specially adapted for this ``needle-in-haystack'' situation.}
\label{fig:tiny}
\end{figure*}

\end{document}